\documentclass[letterpaper, 10 pt, conference]{ieeeconf}  % Comment this line out if you need a4paper

\IEEEoverridecommandlockouts                              % This command is only needed if 
\usepackage{graphicx}
\usepackage{graphics} % for pdf, bitmapped graphics files
\usepackage{epsfig} % for postscript graphics files
\usepackage{amsmath} % assumes amsmath package installed
\usepackage{amssymb}  % assumes amsmath package installed
\usepackage{booktabs}
\usepackage{tabularx}
\usepackage{array}
\usepackage{multicol}
\usepackage{multirow}
\usepackage{float}
\usepackage{array}
\usepackage[utf8]{inputenc}
\usepackage{enumerate} 
\usepackage{caption}
\usepackage{cite}
\usepackage{adjustbox}
\usepackage{color, soul}
\usepackage{amsmath} 
\usepackage{comment}
\usepackage{svg}
\usepackage{xcolor}
\usepackage{diagbox}

\renewcommand{\thesubsection}{\thesection.\Alph{subsection}}

\title{\LARGE \bf
TACS-Graphs: Traversability-Aware Consistent Scene Graphs for Ground Robot Localization and Mapping
}

\definecolor{customGreen}{rgb}{144,238,144}

\author{
Jeewon Kim$^{1}$, Minho Oh$^{1,2}$, and Hyun Myung$^{1*}$, \textit{Senior Member, IEEE} % <-this % stops a space
\thanks{*Corresponding author: Prof. Hyun Myung is with School of Electrical Engineering and KI-R at KAIST.}% <-this % stops a space
\thanks{$^{1}$All authors are with the School of Electrical Engineering, KAIST (Korea Advanced Institute of Science and Technology), Daejeon, 34141, Republic of Korea. \{\tt\footnotesize ddarong2000, minho.oh, hmyung\}@kaist.ac.kr}%
\thanks{$^{2}$URobotics Corp., Republic of Korea. \tt\footnotesize minho.oh@urobotics.ai
}%
\thanks{This work was supported in part by the National Research Foundation of Korea(NRF) grant funded by the Korea government(MSIT) (No. RS-2024-00348461), and in part by the National Research Council of Science \& Technology(NST) grant by the Korea government (MSIT) (No. GTL25041-100). The students are supported by the BK21 FOUR (Republic of Korea).}
}

\begin{document}
\maketitle
% \thispagestyle{empty}
% \pagestyle{empty}

%%%%%%%%%%%%%%%%%%%%%%%%%%%%%%%%%%%%%%%%%%%%%%%%%%%%%%%%%%%%%%%%%%%%%%%%%%%%%%%%
\begin{abstract}
Scene graphs have emerged as a powerful tool for robots, providing a structured representation of spatial and semantic relationships for advanced task planning. Despite their potential, conventional 3D indoor scene graphs face critical limitations, particularly under- and over-segmentation of room layers in structurally complex environments. Under-segmentation misclassifies non-traversable areas as part of a room, often in open spaces, while over-segmentation fragments a single room into overlapping segments in complex environments. These issues stem from naive voxel-based map representations that rely solely on geometric proximity, disregarding the structural constraints of traversable spaces and resulting in inconsistent room layers within scene graphs. 
To the best of our knowledge, this work is the first to tackle segmentation inconsistency as a challenge and address it with \textbf{Traversability-Aware Consistent Scene Graphs~(TACS-Graphs)}, a novel framework that integrates ground robot traversability with room segmentation. By leveraging traversability as a key factor in defining room boundaries, the proposed method achieves a more semantically meaningful and topologically coherent segmentation, effectively mitigating the inaccuracies of voxel-based scene graph approaches in complex environments. Furthermore, the enhanced segmentation consistency improves loop closure detection efficiency in the proposed \textbf{Consistent Scene Graph-leveraging Loop Closure Detection~(CoSG-LCD)} leading to higher pose estimation accuracy. Experimental results confirm that the proposed approach outperforms state-of-the-art methods in terms of scene graph consistency and pose graph optimization performance.
\end{abstract}

%%%%%%%%%%%%%%%%%%%%%%%%%%%%%%%%%%%%%%%%%%%%%%%%%%%%%%%%%%%%%%%%%%%%%%%%%%%%%%%%

\vspace{0.5em}
\section{Introduction}
\label{sec:intro}

In recent years, scene graphs have become a powerful tool for representing and interpreting 3D spaces in robotics~\cite{rosinol20203d,rafia2018sgwarehouse,wang2023rs2g,zhang2024adinteractionscenegraph,lv2024t2sg,jiao2022sequentialmanipulationplanningscene,ni2024grid}. By capturing semantic and spatial relationships between objects and spaces, scene graphs provide a comprehensive understanding of complex environments. Scene graphs have been leveraged for diverse applications, including optimizing navigation and task sequences in logistics and autonomous driving~\cite{rafia2018sgwarehouse,wang2023rs2g,zhang2024adinteractionscenegraph,lv2024t2sg}, facilitating efficient task planning. Additionally, they enhance human-robot interaction, such as IoT-based communication, by offering intuitive interfaces that bridge technology and human intent~\cite{jiao2022sequentialmanipulationplanningscene,ni2024grid}. Building on these capabilities, scene graphs enable robots to adapt to their surroundings, supporting real-time decision-making and precise task execution in various scenarios~\cite{agia2022taskography}.

Despite their effectiveness, most scene graphs used in previous studies were limited to 2D representations, defining spatial relationships in simplistic terms such as ``in front of'' or ``next to.'' While sufficient for simple tasks, these methods struggle to capture the complexity of real-world 3D spatial relationships. To overcome this limitation, 3D scene graphs have been developed, allowing more accurate spatial and semantic modeling. 
Recent 3D scene graph-based approaches often incorporate room-level task planning, providing a structured framework that enhances robot interaction, spatial understanding, and navigation efficiency~\cite{ni2024grid,agia2022taskography,DBLP,hov-sg}. Expanding on this foundation, recent studies have explored improving mapping accuracy through scene graphs. Kimera~\cite{rosinol2021kimera} and Hydra~\cite{hughes2022hydra} utilize RGB-D cameras for dynamic 3D scene graphs. However, vision-based pose estimation is susceptible to environmental conditions, and pose graph optimization (PGO) in previous methods is triggered only upon loop closure, limiting real-time adaptability. To address these challenges, Bavle~\textit{et al.} introduced LiDAR-based approaches~\cite{bavle2022situational,bavle2023s,millan2023better}, where a hierarchical factor graph is tightly integrated with simultaneous localization and mapping~(SLAM), improving pose optimization.

Even with these advancements, existing 3D scene graphs still face significant challenges in complex environments, particularly regarding under-segmentation and over-segmentation in room layers. Since room segmentation is closely tied to free-space segmentation, these issues stem from the limitations of voxel-based 3D mapping techniques like Octomap~\cite{octomap} and Euclidean signed distance fields~(ESDFs)~\cite{esdf}, which are widely used in scene graph generation. 
Under-segmentation occurs because voxel-based room segmentation methods rely on environmental topology~\cite{hughes2022hydra}, causing non-traversable areas, such as open spaces beyond railings, to be misclassified as part of a room. 
Over-segmentation, on the other hand, occurs when high computational demands disrupt spatial consistency, causing a single room to be fragmented into multiple overlapping segments. This issue is exacerbated in large or complex environments, where voxel-based methods fail to maintain continuity, leading to disjoined spaces. %Additionally, overlapping free-space regions between adjacent rooms complicate scene graph generation. As the computational burden increases, real-time updates become less reliable, further exacerbating segmentation issues. (image) 
These segmentation issues induce inconsistencies in the scene graphs, hindering consistent representation of the same location across trials and causing variations over time. Consequently, reliability and usability of scene graphs for task planning are compromised.

To address this inconsistency, we propose incorporating ground robot traversability as a key factor in room segmentation, replacing conventional voxel-based free-space representations. Unlike voxel-based methods that segment spaces based solely on topology, our approach integrates the robot’s ability to traverse through the environment, ensuring more consistent and semantically meaningful segmentation. To the best of our knowledge, this is the first study to introduce a traversability-aware framework for scene graph generation.

In summary, the contributions of this paper address the following key challenges in 3D scene graph generation:
\begin{enumerate}[1)] 
    \item \textbf{Under- and over-segmentation of rooms.} We propose \mbox{\textit{TACS-Graphs}}, a scene graph framework that leverages ground robot traversability for room segmentation, ensuring consistent and semantically meaningful spatial representations while addressing under- and over-segmentation issues in voxel-based methods.
    \item \textbf{Loop closure redundancy and distance threshold dependency.} We propose \mbox{\textit{CoSG-LCD}}, a loop closure detection module that enhances PGO by leveraging the consistent room layer of scene graphs, reducing unnecessary loop closures and mitigating dependency on fixed distance thresholds for loop detection.
    \item \textbf{Scene graph dependency on precise pose estimation.} We establish interdependency between scene graph consistency and PGO, demonstrating that consistent scene graphs improve loop closure, while refined robot poses further enhance graph consistency.
\end{enumerate}
\section{Related Works}
\label{sec:related_works}

\subsection{Scene Graphs}
Scene graphs are structured representations that capture spatial and semantic relationships within and between objects and places, enabling the hierarchical composition of simple relationships to explain complex relationships. Unlike traditional geometry-focused mapping, scene graphs integrate geometric and semantic interrelations, providing a deeper contextual understanding of the environments~\cite{LI2024127052}. This integration enables both robot-environment and robot-human interaction by structuring spatial information in an interpretable manner, enabling intelligent operations and improving real-time decision-making. Widely adopted in applications such as warehouse automation and urban navigation, scene graphs optimize path planning and task execution~\cite{rafia2018sgwarehouse,wang2023rs2g,zhang2024adinteractionscenegraph,lv2024t2sg}. Therefore, given their role in robotic perception and interaction, real-time scene graph generation with human-intuitive representations is essential for efficient operation in complex environments.
%Scene graphs act as user-friendly interfaces, connecting robots with humans and their surroundings. They enhance real-time 3D environmental understanding, simplify interaction with intuitive commands, and optimize tasks such as path planning in warehouses or urban settings~\cite{rafia2018sgwarehouse,wang2023rs2g,zhang2024adinteractionscenegraph,lv2024t2sg}. Therefore, the ability of scene graph generation in real-time with human-familiar representation is crucial for enhancing robot task performance.

While early studies primarily focused on 2D scene graphs for computer vision applications, their flat representations struggled to capture the complex spatial relationships of real-world environments. To address this limitation, 3D scene graphs were introduced, directly modeling geometric-semantic relationships on a 3D map for more structured and interpretable representation. Vision-based 3D scene graphs, such as Kimera~\cite{rosinol2021kimera} and Hydra~\cite{hughes2022hydra}, use RGB-D cameras for mesh mapping and object semantic segmentation in dynamic environments. However, they were susceptible to lighting variations and occlusions, which degrade visual SLAM performance. 
To overcome these limitations, LiDAR-based 3D scene graphs, including \mbox{S-Graphs}~\cite{bavle2022situational} and \mbox{S-Graphs+}~\cite{bavle2023s}, were introduced. Unlike previous scene graph generation methods that were not tightly coupled with SLAM, \mbox{S-Graphs} and \mbox{S-Graphs+} use factor graph optimization~(FGO) across hierarchical layers, including keyframe layers, enabling pose graph optimization without relying on loop closure. They demonstrated improved SLAM accuracy by incorporating voxel-based free-space clustering for room segmentation. To enhance room segmentation using only wall information, without relying on free space, Bavle~\textit{et al.} also utilized graph neural networks~(GNNs)~\cite{millan2023better}. However, the GNN-based method struggled in complex and untrained environments. This paper focuses on overcoming the limitations of voxel-based free-space segmentation methods, particularly in large-scale, structurally complex indoor environments.

\subsection{Voxel-based Free-space Segmentation}
With the growing demand for room-wise task execution, precise room segmentation has become a critical component of 3D scene graphs. Voxel-based free-space segmentation and obstacle detection methods are widely employed in existing scene graph frameworks to enhance spatial understanding. 
Among these methods, signed distance fields~(SDFs) are commonly used for 3D mapping. Compared with ray-casting-based approaches like Octomap~\cite{octomap}, SDFs demonstrate superior performance in handling noisy sensor measurements~\cite{oleynikova2016signed}. Truncated SDFs~(TSDFs) further improved computational efficiency by truncating distance calculations, producing smooth surface meshes. However, this truncation compromises accuracy beyond the defined range. To mitigate this limitation, Oleynikova~\textit{et al.}~\cite{esdf} introduced an incremental approach, ESDFs. Utilizing efficient spatial hashing and wavefront propagation, ESDFs extend TSDFs to provide comprehensive Euclidean distance information, facilitating real-time mapping.

However, voxel-based map representations still incur high computational costs in large-scale environments, posing challenges for real-time applications. Moreover, voxel-based free-space segmentation relies solely on environmental topology, limiting its ability to segment semantically distinct rooms in open floor plan environments, as noted in Hydra~\cite{hughes2022hydra}.

\subsection{Traversability Mapping}
\label{sec:traversability_mapping}
Traversability mapping is essential for the safe exploration and efficient navigation of autonomous ground robots. It supports key robotic functions such as obstacle recognition~\cite{zermas2017fast, oh2022travel}, static map generation~\cite{lim2021ral, jang23toss}, odometry estimation~\cite{shan2018lego, song23bigstep}, and global localization~\cite{lim2023quatro++}.

Zermas~\textit{et al.}~\cite{zermas2017fast} proposed a fast ground plane fitting method for ground robot traversal, using principal component analysis to extract initial seeds for ground plane estimation. However, traditional ground segmentation methods often rely on an iterative random sample consensus~(RANSAC) process for each point cloud input, which is computationally expensive and less effective in complex terrains. 
To address these limitations, Oh \textit{et al.}~\cite{oh2022travel} introduced an efficient ground segmentation approach that assumes the lowest elevation points in the point cloud to represent the ground surface. This method iteratively refines ground plane estimation, classifies ground points, and finds traversable areas, achieving both computational efficiency and precise representation in diverse environments. Building on \cite{oh2022travel}, \textit{B-TMS}~\cite{oh2024btms} further enhanced traversability mapping by inferring occluded regions using a Bayesian generalized kernel~(BGK), improving performance in environments with significant occlusions.

Beyond navigation, a traversability map provides a structured approach for efficiently clustering free-space regions suitable for movement by segmenting traversable areas. This capability is particularly valuable in scene graph generation, where structural segmentation aligned with human perception and rapid mapping is essential. This paper proposes a pioneering method that integrates ground robot traversability mapping into scene graph generation. By leveraging terrain continuity with traversability constraints, our approach accounts for discontinuities such as stairs and gaps, enhancing room segmentation consistency. This results in a structured representation that aligns more closely with human perception, overcomes the limitations of traditional voxel-based free-space methods, and improves overall scene graph consistency in complex indoor environments.  

\section{Traversability-Aware Consistent Scene Graphs}
\label{sec:approach}

\begin{figure}[t!]
    \centering
    \includegraphics[width=\linewidth]{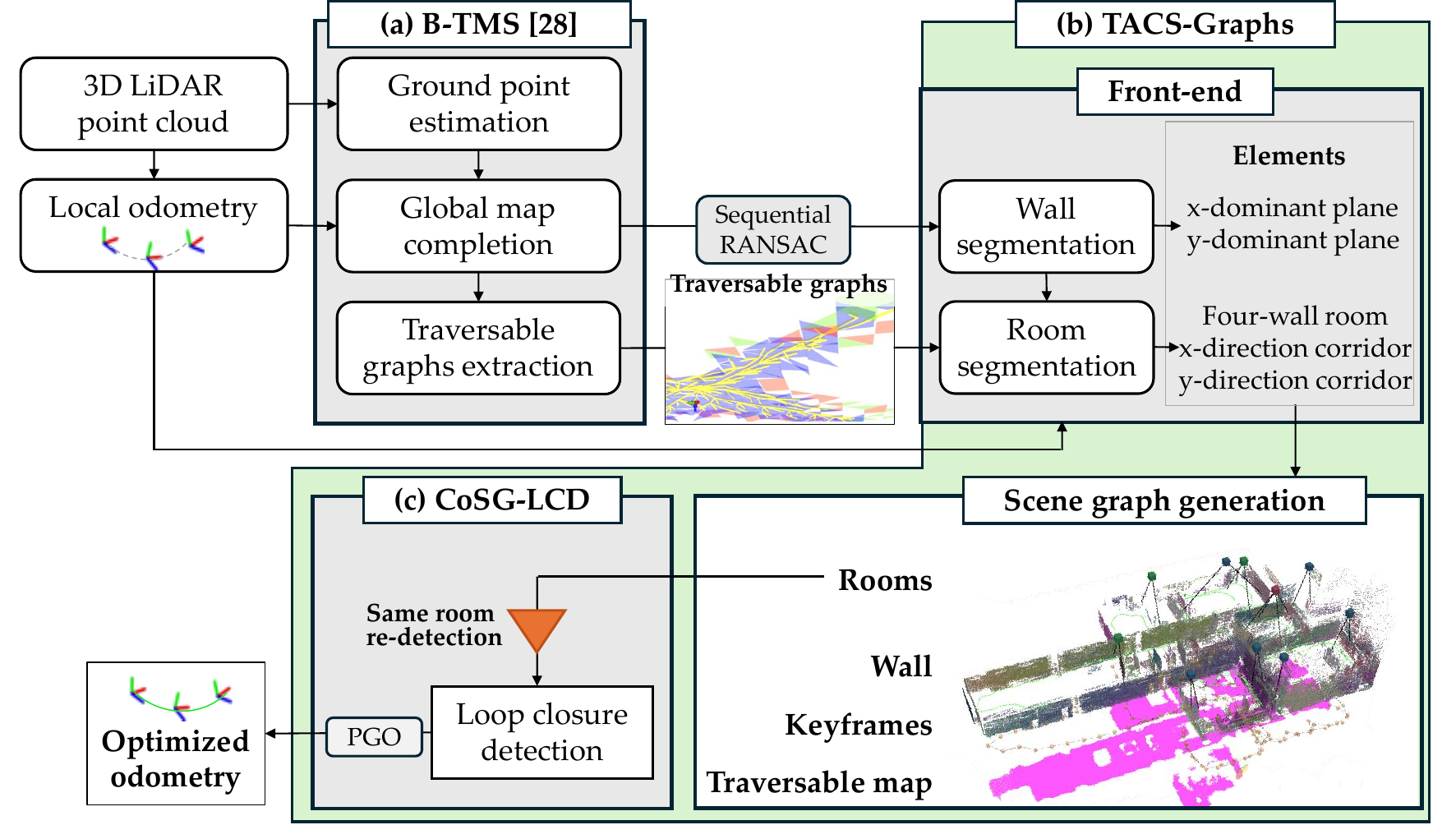}
    \caption{An overview of \mbox{\textit{TACS-Graphs}}, the proposed traversability-aware scene graph generation framework for ground robots. (a)~A real-time traversable graph is extracted using B-TMS~\cite{oh2024btms}. (b)~Room segmentation is performed using traversable graphs and walls, resulting in consistent scene graphs. (c)~This consistency enhances \mbox{\textit{CoSG-LCD}} module, a loop closure detection method that recognizes revisited rooms.}
    \label{fig:framework}
\end{figure}

% voxel based free-space clustering에 초점맞춰서 proposed가 차이 보이는 이유 설명 보충
While previous 3D scene graph generation methods perform well in simple environments, they struggle to maintain consistency in complex settings. The proposed \mbox{\textit{TACS-Graphs}} addresses this limitation by incorporating traversability-aware room segmentation. The overall process, illustrated in Fig.\ref{fig:framework}, consists of the following steps.

First, the scene graph uses LiDAR-based local odometry to estimate the robot’s motion and extracts traversable graphs from 3D LiDAR scans using B-TMS~\cite{oh2024btms}~(\mbox{Section~\ref{sec:btms}}). Traversable rooms are then segmented through wall detection and traversability-aware free-space clustering~(\mbox{Section~\ref{sec:room_seg}}). To further prevent over-segmentation, rooms with overlapping free spaces are merged~(\mbox{Section~\ref{sec:rm}}), ensuring spatial continuity and minimizing redundant partitions. As a result, the proposed method prevents segmentation issues and maintains topological integrity, improving consistency in the \texttt{Wall} and \texttt{Room} layers. Leveraging this consistent graph, the proposed \mbox{\textit{CoSG-LCD}} module efficiently detects loop closures by identifying optimal loop pairs in revisited rooms~(\mbox{Section~\ref{sec:loop_detection}}). Consequently, the corrected odometry and optimized scene graph iteratively refine each other, enhancing both localization accuracy and scene graph consistency.

% \begin{itemize}
%     \item \textbf{Traversable Graph Extraction:} Following the B-TMS ground segmentation method~\cite{oh2024btms}, a traversable graph is extracted from the ground in real-time.
%     \item \textbf{Traversability-Aware Room Segmentation:} By accumulating traversable graphs over time until no further new graphs are generated (e.g., when the robot exits the room), room segmentation is performed using the traversable graphs and extracted walls.
%     \item \textbf{Room Merging (RM) Module:} Rooms with overlapping traversable free-space are merged to address over-segmentation.
%     \item \textbf{Efficient Loop Detection:} Consistent room segmentation enhances loop closure detection by identifying re-detected rooms as identical, limiting keyframe searches to neighboring areas. This improves computational efficiency and strengthens scene graph consistency through accurate pose updates.
% \end{itemize} 

\subsection{Traversable Graph Extraction}
\label{sec:btms}
B-TMS~\cite{oh2024btms} is used as the traversability mapping technique in this paper. This method segments 3D point cloud maps into triangular grids and identifies traversable ground and obstacles. It quantifies the traversability of ground points using BGK inference to account for occluded and undetected regions. A traversable graph is then constructed from the robot’s position using a breadth-first search algorithm, where triangular grid nodes are connected by edges linking neighboring grids.  The graph is updated in real-time, enabling efficient navigation in complex environments.

\subsection{Room Segmentation}
\label{sec:room_seg}
\subsubsection{Wall Layer Segmentation}
Walls are extracted using a sequential RANSAC algorithm, which iteratively detects dominant large planar surfaces in 3D point clouds~\cite{bavle2022situational}. Detected vertical planes in the robot's body frame, represented as individual planes \mbox{${}^B{\pi} \in {\mathbb{R}}^3$} within the plane set \mbox{${}^B{\Pi}$}, are transformed into the map frame as \mbox{${}^M{\pi} \in {\mathbb{R}}^3$} within the corresponding set \mbox{${}^M{\Pi}$}. Walls are categorized into \textit{x}-direction planes (${{}^M\Pi_x}$) and \textit{y}-direction planes (${{}^M\Pi_y}$) based on their dominant normal components, ${}^M{n_x}$ and ${}^M{n_y}$, respectively. Each plane is further classified based on the sign of its normal component to determine the wall's facing direction in the room: ${}^M\Pi_{x_+}$ for \mbox{${}^B{n_x}\ge0$}, ${}^M\Pi_{x_-}$ for \mbox{${}^B{n_x}<0$}, ${}^M\Pi_{y_+}$ for \mbox{${}^B{n_y}\ge0$}, and ${}^M\Pi_{y_-}$ for \mbox{${}^B{n_y}<0$}, to determine the wall's facing direction in the room.
%To determine the wall's facing direction in the room, each plane is further classified based on the sign of its dominant normal component in the robot body frame. The \textit{x}-direction planes are divided into ${}^M\Pi_{x_+}$ for \mbox{${}^B{n_x}\ge0$} and ${}^M\Pi_{x_-}$ for \mbox{${}^B{n_x}<0$}, while the \textit{y}-direction planes are classified as ${}^M\Pi_{y_+}$ for \mbox{${}^B{n_y}\ge0$} and ${}^M\Pi_{y_-}$ for \mbox{${}^B{n_y}<0$}.

Room segmentation is performed by identifying walls adjacent to clustered free-space within the room. We exploited the Manhattan world assumption, where corridors are defined by two parallel walls, and enclosed four-wall rooms are surrounded by two pairs of parallel walls.

\subsubsection{Free-space Segmentation}
\begin{figure}[t!]
    \centering
    \includegraphics[width=0.9\linewidth]{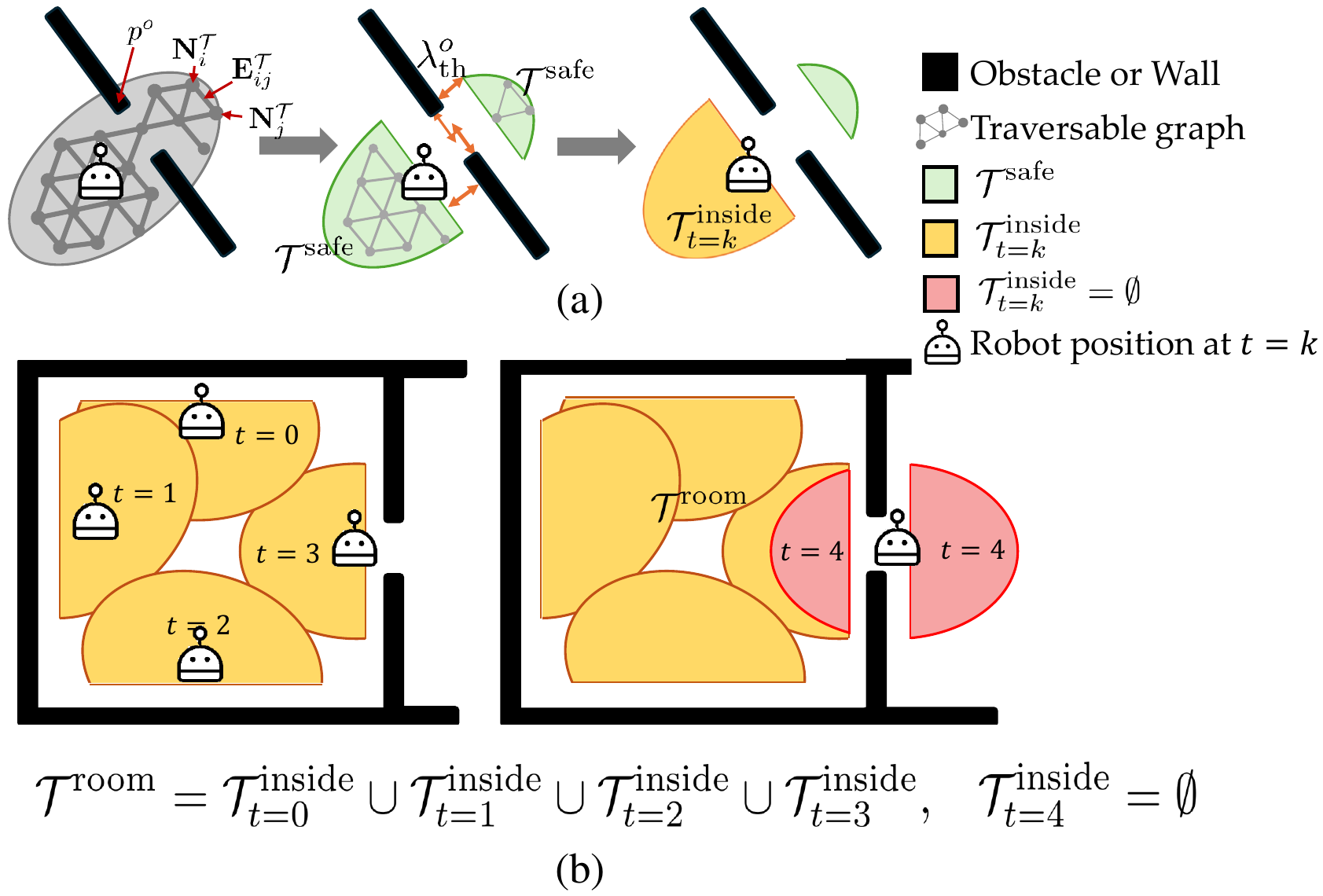}
    \caption{An illustration of free-space clustering and room segmentation using traversable graphs. (a)~Process of obtaining $\mathcal{T}_k^{\text{inside}}$ at time \mbox{$t=k$}. (b)~An example where $\mathcal{T}^{\text{room}}$ is incrementally formed. Once no new $\mathcal{T}^{\text{inside}}$ is detected at $t=4$ as the robot exits a room, the accumulated $\mathcal{T}^{\text{room}}$ from \mbox{$t=0$} to \mbox{$t=3$} is used to define room walls.}
    \label{fig:freespace}
\end{figure}

At time \mbox{$t = k$}, free-space clusters of the current room are segmented using the traversable graph from \mbox{Section~\ref{sec:btms}}, as shown in \mbox{Fig.~\ref{fig:freespace}(a)}. To distinguish inside-room and outside-room regions, nodes near obstacles are disconnected based on a distance threshold $\lambda_{\text{th}}^o$. Thus, nodes near doorways are separated when the robot exits the room, forming distinct subgraphs for interior and exterior spaces. Over time, the robot constructs a safe graph \( \mathcal{T}^{\text{safe}} \), which consists of the node set \( \mathcal{N}^{\text{safe}} \) and the edge set \( \mathcal{E}^{\text{safe}} \), defined as $\mathcal{T}^{\text{safe}} = (\mathcal{N}^{\text{safe}}, \mathcal{E}^{\text{safe}})$ where
\begin{equation}
    \begin{split}
    \mathcal{N}^{\text{safe}} = \left\{ \mathbf{N}^{\mathcal{T}}_i \mid \min_{\mathbf{p}^o \in \mathbf{P}^o} d(\mathbf{N}^{\mathcal{T}}_i, \mathbf{p}^o) \geq \lambda_{\text{th}}^o \right\} \\
    \mathcal{E}^{\text{safe}} = \big\{ (\mathbf{N}^{\mathcal{T}}_i, \mathbf{N}^{\mathcal{T}}_j) 
    \mid \ \mathbf{N}^{\mathcal{T}}_i, \mathbf{N}^{\mathcal{T}}_j \in \mathcal{N}^{\text{safe}}
    \big\}.
    \end{split}
\end{equation}
Here, the node set includes all traversable nodes \( \mathbf{N}^{\mathcal{T}}_i \) that satisfy the safe distance condition from the nearest obstacle position \mbox{$\mathbf{p}^o \in \mathbf{P}^o$}, where \( d(\cdot) \) denotes the Euclidean distance. The edge set consists of pairs of nodes \mbox{$(\mathbf{N}^{\mathcal{T}}_i, \mathbf{N}^{\mathcal{T}}_j)$} that belong to \( \mathcal{N}^{\text{safe}} \). This formulation ensures that the resulting graph excludes regions near obstacles and outside doorways, effectively segmenting the free space within the current room,~\( \mathcal{T}_{t = k}^{\text{inside}} \), at time \( t = k \).

% \begin{equation} 
%     \begin{split}
%         \mathcal{T}^{\text{safe}} = 
%         \left\{ \mathbf{N}^{\mathcal{T}}_i \mid \min_{\mathbf{p}^o \in \mathbf{P}^o} d(\mathbf{N}^{\mathcal{T}}_i, \mathbf{p}^o) \geq {\lambda_{\text{th}}^o} \right\} \\
%         \cup \left\{ \mathbf{E}^{\mathcal{T}}_{ij} \mid \mathbf{N}^{\mathcal{T}}_i \in \mathcal{T}^{\text{safe}} \text{ and } \mathbf{N}^{\mathcal{T}}_j \in \mathcal{T}^{\text{safe}} \right\},
%     \end{split}
%     \label{eq:graph_th} 
% \end{equation}
% where $d(\cdot)$ represents the Euclidean distance, \mbox{$\mathbf{P}^o = \{ \mathbf{p}^o \mid \mathbf{p}^o \in \mathbb{R}^3 \}$} denotes the set of obstacle positions, and \mbox{$\mathbf{N}^{\mathcal{T}}_i \in \mathbb{R}^3$} represents the set of traversable nodes satisfying the safe distance condition, forming $\mathcal{T}^{\text{safe}}$. The edge \mbox{$\mathbf{E}^{\mathcal{T}}_{ij} \in \mathbb{R}^3$} connects the nodes $\mathbf{N}^{\mathcal{T}}_i$ and $\mathbf{N}^{\mathcal{T}}_j$. This formulation ensures that the resulting graph excludes regions outside the doorways or near obstacles, effectively segmenting the free-space within the robot's current room, $\mathcal{T}_k^{\text{inside}}$, at time $t=k$.

\subsubsection{Room Extraction}

\mbox{Fig.~\ref{fig:freespace}(b)} illustrates the incremental formation of a room-level traversable graph, $\mathcal{T}^{\text{room}}$, which refines room boundaries and finalizes segmentation. As the robot moves through rooms or corridors, traversable subgraphs,~$\mathcal{T}_{t=k}^{\text{inside}}$, accumulate over time from \mbox{$t=0$} to \mbox{$t=T$}. When the robot reaches a doorway or a narrow passage at \mbox{$t=T+1$}, the traversable graph splits at the robot's position, and no further connected graphs exist in its vicinity. The final room-level traversable graph, $\mathcal{T}^{\text{room}}$, is then obtained as the union of all accumulated $\mathcal{T}_{t=k}^{\text{inside}}$, formulated as follows:
\begin{equation} 
    \mathcal{T}^{\text{room}} = \bigcup_{t=0}^{T} \mathcal{T}_{t}^{\text{inside}}, \quad \mathcal{T}_{t=T+1}^{\text{inside}} = \emptyset.
    \label{eq:room_union} 
\end{equation}

To segment a room using the walls and $\mathcal{T}^{\text{room}}$, the walls with the largest number of adjacent points to $\mathcal{T}^{\text{room}}$ are selected from each wall category: ${}^M\Pi_{x_+}$, ${}^M\Pi_{x_-}$, ${}^M\Pi_{y_+}$, and ${}^M\Pi_{y_-}$. If two parallel walls contain a sufficient number of adjacent points, the room is classified as an ``\textit{x}-direction'' or ``\textit{y}-direction'' corridor. If two pairs of walls in both the x and y directions satisfy the adjacency condition, the room is classified as a ``four-wall room.'' Each identified room is represented as a node in the scene graph, with its geometric center, determined as the midpoint between the bounding walls, serving as the node position. The segmented room node and its corresponding walls are stored in the \texttt{Room} and \texttt{Wall} layers, respectively.

%%%%%%%%%%%%%%%%%%%%%%%%%%%%%%%%%%%%%%%%%%%%%%%%%%%%%%%%%%%%%%%

\subsection{Room Merging Module}
\label{sec:rm}
Existing scene graph generation methods do not update room nodes upon room revisits~\cite{hughes2022hydra,rosinol2021kimera}. S-Graphs+~\cite{bavle2023s} addresses this by connecting room nodes when their Euclidean distance satisfies \mbox{$d(x,y)\leq 1~\mathrm{m}$} and leveraging these connections for additional PGO through factor graph optimization. However, this fixed threshold often results in over-segmentation in complex or large environments.

\begin{figure}[t!]
    \centering 
    \includegraphics[width=\linewidth]{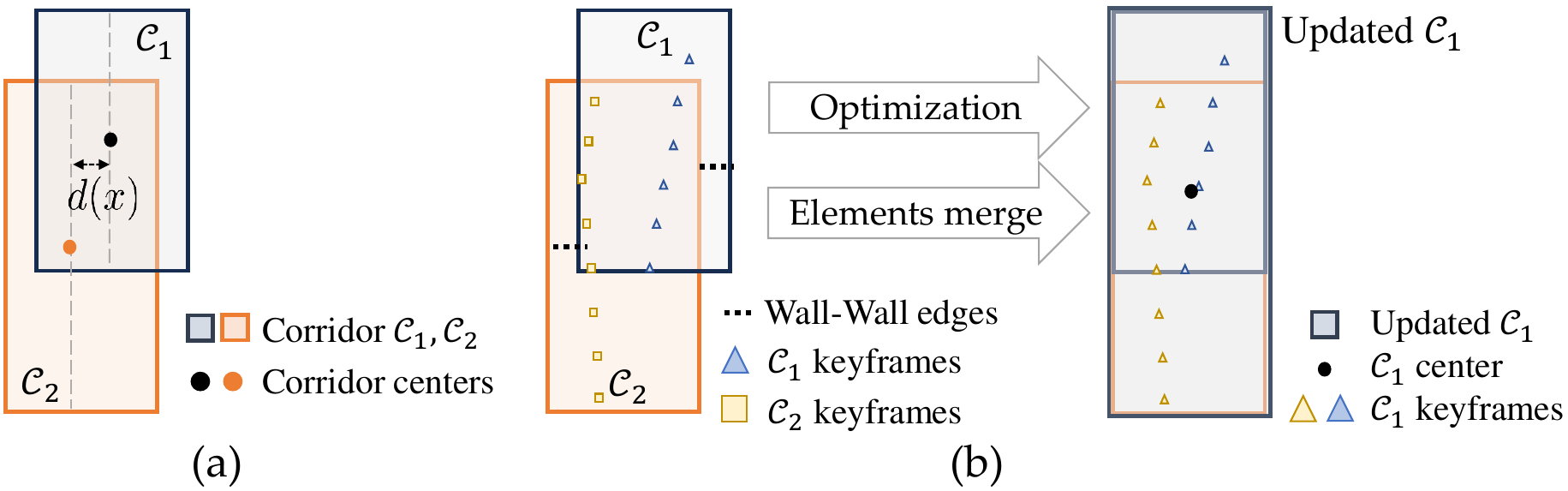}
    \caption{An example of the proposed room merging module. (a)~Before merging, two \textit{y}-direction corridors \(\mathcal{C}_1\) and \(\mathcal{C}_2\) are distinct, with overlapping traversable graphs and a small center distance \(d(x)\). (b)~After optimizing keyframes and the center of \(\mathcal{C}_2\) to make the wall-to-wall edge converge to zero, they merge, updating \(\mathcal{C}_1\) as a unified room.} 
    \label{fig:roommerge} 
\end{figure}

The proposed room merging~(RM) module improves consistency by merging rooms with overlapping traversable free-space while evaluating centroid similarity along orthogonal axes. For instance, as illustrated in \mbox{Fig.~\ref{fig:roommerge}(a)}, two \textit{y}-direction corridors detected at similar locations are compared based on their \textit{x}-axis distance. This distance is given by \mbox{$d(x)=\lvert x_{{\kappa}}-x_{\tilde{\kappa}} \rvert$}, where $\kappa$ represents an existing \textit{y}-direction corridor node ($\mathcal{C}_1$), and $\tilde{\kappa}$ denotes a newly detected corridor node ($\mathcal{C}_2$). Throughout this section, the tilde notation ($\tilde{\cdot}$) is used to indicate newly detected room components, distinguishing them from previously existing structures. The wall-to-wall edges are then generated using the following cost function $c_{\kappa}(\cdot)$:
\begin{equation}
    \begin{split}
    c_{\kappa}(\left[{\pi}_{x_+},{\pi}_{x_-},{\kappa}\right])&
    = \left\|{f_x}({\tilde{\pi}}_{x_{+}},{\tilde{\pi}}_{x_{-}},\tilde{\kappa})-d(x)\right\|^2,
    \end{split}
\label{eq:residual}
\end{equation}
where \mbox{$\left[{\pi}_{x_+},{\pi}_{x_-}\right]$} is the pair of walls forming the node $\kappa$. The function ${f_x}(\cdot)$ models all \textit{x}-axis components of the new room node, which has a center position $\tilde{\kappa}$, and walls $\tilde{\pi}_{x_+}$ and $\tilde{\pi}_{x_-}$.

After performing PGO, the position of the node $\kappa$ is updated by integrating the components of the new room, including free-space and keyframes, as shown in \mbox{Fig.~\ref{fig:roommerge}(b)}. This room merging strategy mitigates over-segmentation issues in complex environments, thereby improving the semantic consistency of the scene graph.

%%%%%%%%%%%%%%%%%%%%%%%%%%%%%%%%%%%%%%%%%%%%%%%%%%%%%%%%%%%%%%%
\definecolor{customOrange}{rgb}{1.0,0.4,0.3}
\definecolor{customGreen}{rgb}{0.3,0.8,0.3}

\subsection{Consistent Scene Graph-based Loop Detection}
\label{sec:loop_detection}

When a robot pose is estimated using local odometry, accumulated drift causes global inconsistency, leading to variations in room segmentation even for the same locations. Because these inconsistencies propagate to scene graph generation, accurate localization is essential for maintaining graph consistency.
%When a robot pose is estimated using local odometry, the estimated pose is not globally consistent due to accumulated local drifts. This global inconsistency in pose can lead to different room segmentation results even for the same rooms, inducing inconsistencies in scene graph generation. Thus, loop closure is necessary for scene graph generation, which relies on precise localization. 
%Studies like Qian \textit{et al.}~\cite{qian2022loop} and Hydra~\cite{hughes2022hydra} have improved loop closure accuracy using semantic ships and object associations, but object-layer-based methods often fail in texture-sparse or complex environments. Additionally,

\begin{figure}[t!]
    \centering 
    \includegraphics[width=0.8\linewidth]{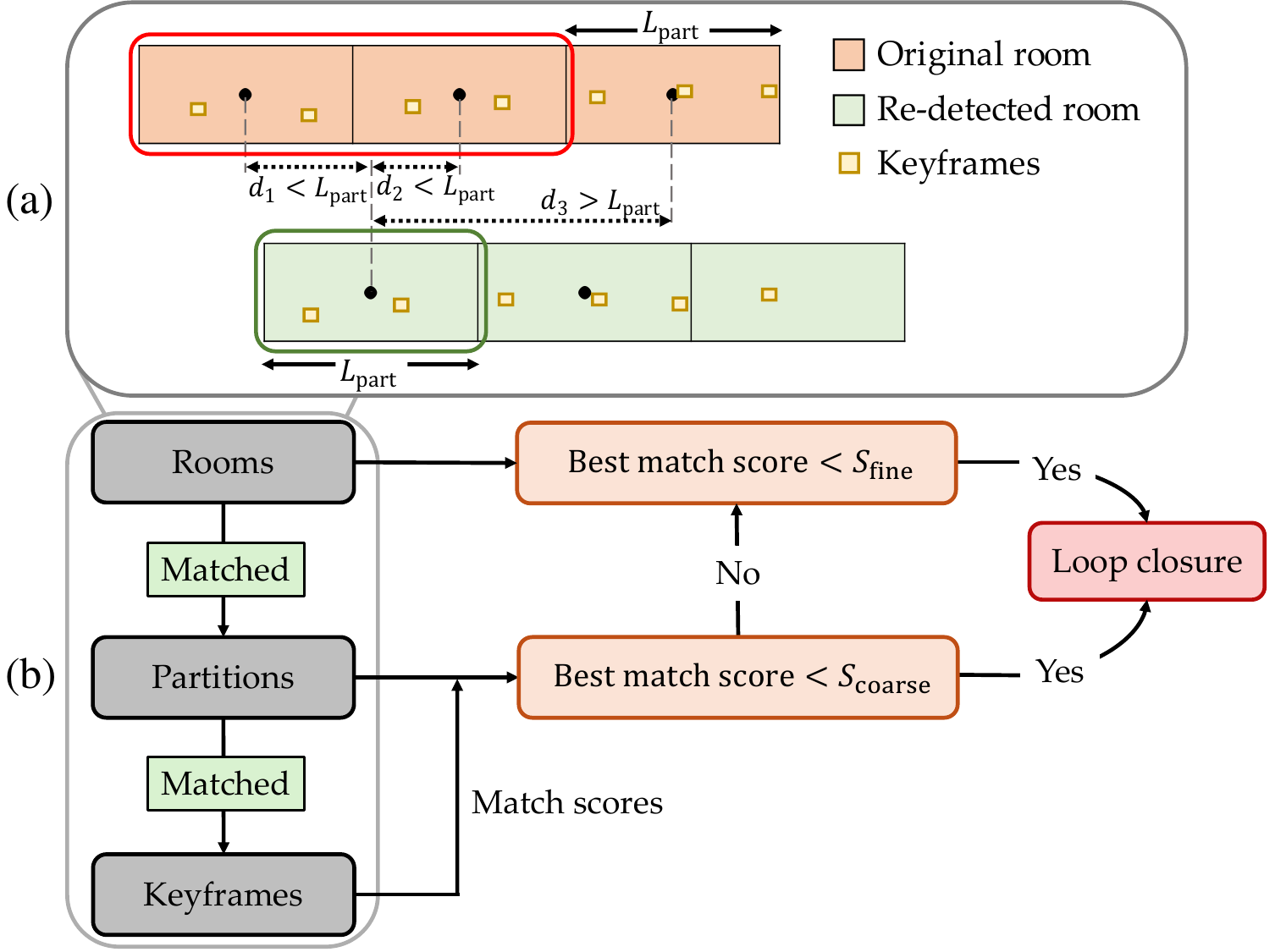}
    \caption{Overview of the proposed \mbox{\textit{CoSG-LCD}} loop closure detection framework. (a)~Loop closure detection with a coarse-to-fine strategy. If no keyframe pair exceeds \(S_{\text{coarse}}\), a relaxed threshold \(S_{\text{fine}}\) is applied to ensure robustness against pose estimation errors. (b)~For large rooms, partitions of length \(L_{\text{part}}\) are created, and the best keyframe pair in each partition is selected for loop closure if its match score exceeds \(S_{\text{coarse}}\).} 
    \label{fig:loop_framework} 
\end{figure}

However, traditional scene graphs typically rely on repetitive loop closure processes, which are both time-inefficient and prone to failure when initial pose estimates are inaccurate. To address this issue, the proposed \mbox{\textit{CoSG-LCD}} method triggers loop closure through same-room re-detection, avoiding unnecessary and redundant loop closures. This method is illustrated in Fig.~\ref{fig:loop_framework}. 
The method leverages the consistent room layer of the scene graph obtained from~\mbox{Section~\ref{sec:room_seg}}, increasing the likelihood of recognizing previously visited rooms. 
To reduce computational overhead, keyframe candidates are restricted to those from the same room detected in earlier frames. While a small distance threshold for keyframe searching for loop detection~($d^{\text{loop}}_{\text{th}}$) works well with minimal initial pose errors, it may fail when such errors are large. Therefore, a sufficiently large \mbox{$d^{\text{loop}}_{\text{th}}=5~\mathrm{m}$} is used in the proposed loop closure detection module to ensure robustness. 
Additionally, for large rooms, imaginary partitions of length $L_{\text{part}}$ are introduced to constrain keyframe searches to neighboring regions, as shown in \mbox{Fig.~\ref{fig:loop_framework}(a)}. When a previously visited room~(\textcolor{customOrange}{orange}) is re-detected~(\textcolor{customGreen}{green}), keyframe matching is conducted within the two nearest partitions of the original room, where the distances between partition centers satisfy \mbox{$d_1, d_2<L_{\text{part}}$}. 
Then, the keyframe selection process for coarse-to-fine loop closure detection is illustrated in \mbox{Fig.~\ref{fig:loop_framework}(b)}. The best match is selected among the keyframes if its score exceeds a threshold,~$S_{\text{coarse}}$, triggering loop closure with the matched keyframe pair. However, when initial pose estimation is highly inaccurate, keyframe candidates may fail to meet this criterion. To mitigate this, a relaxed matching threshold,~$S_{\text{fine}}$, is introduced to allow weaker initial matches. This enables a coarse-to-fine refinement process that improves loop closure robustness despite initial pose inaccuracies.
\section{Experiments and Evaluation}
\label{sec:evaluation}
To evaluate the effectiveness of our method, we conducted experiments to address the following key questions:
\begin{enumerate}
    \item \textbf{Under- and over-segmentation of rooms.} Does the proposed method mitigate these issues and improve scene graph consistency?~(Section~\ref{sec:consistency_eval})
    \item \textbf{Loop closure redundancy and distance threshold dependency.} Does the proposed method enhance PGO by reducing redundant loop closures and mitigating dependency on fixed distance thresholds?~(Section~\ref{sec:pgo_perf})
    \item \textbf{Scene graph dependency on precise pose estimation.} Does the proposed method overcome the impact of inaccurate localization on graph consistency, and improve PGO performance when dealing with inconsistent graphs?~(Section~\ref{sec:ablation})
\end{enumerate}

\subsection{Experimental Setup}
For comparison of scene graph-based SLAM methods, including our proposed approach, Fast-GICP~\cite{koide2020fastgicp} was used for local pose estimation. Loop closure was handled using normal distribution transform-based scan matching~\cite{hdlslam}, which identifies candidates through translational thresholding between robot poses. All experiments were conducted on a workstation with an Intel Core i9-12900K CPU, 32 GB of RAM, and an NVIDIA GeForce RTX 3070 GPU. %We compared our approach with scene graph generation methods incorporating PGO, and state-of-the-art LiDAR-based SLAM methods using scan context-based backends.

\subsection{Dataset}
\label{sec:dataset}
\subsubsection{TIERS LiDAR Dataset}  
\begin{figure}[b]
    \centering
    \includegraphics[width=\linewidth]{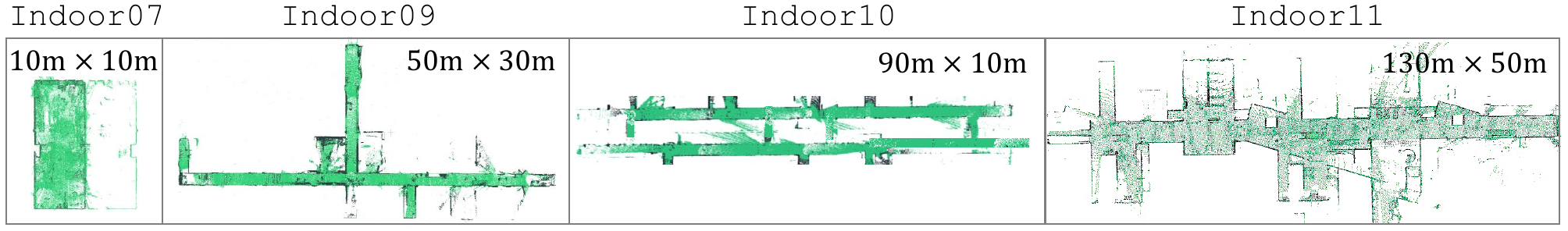}
    \caption{Layouts of four TIERS indoor datasets~\cite{tiers2023benchmark}. From left to right: \texttt{Indoor07}, \texttt{Indoor09}, \texttt{Indoor10}, and \texttt{Indoor11}.}
    \label{fig:tiers}
\end{figure}

We selected four sequences from the TIERS LiDAR dataset~\cite{tiers2023benchmark} to quantitatively assess graph consistency and PGO performance. These indoor datasets feature looped robot trajectories (Fig.~\ref{fig:tiers}): \texttt{Indoor07} contains repetitive loops in a single room, \texttt{Indoor09} includes a squared corridor, and \texttt{Indoor10} and \texttt{Indoor11} span multiple rooms and corridors. The dataset provides sensor data from various LiDARs, RGB cameras, and inertial measurement units~(IMUs), along with ground truth odometry.

\subsubsection{S-Graphs Dataset}
To validate performance in a simple environment, \mbox{S-Graphs} dataset~\cite{bavle2022situational} is used, collected at an indoor construction site. A quadruped robot equipped with a LiDAR sensor explored structured small rooms, capturing multiple loops.

\subsubsection{Real-world Dataset}  
\begin{figure}[b]
    \centering
    \includegraphics[width=0.88\linewidth]{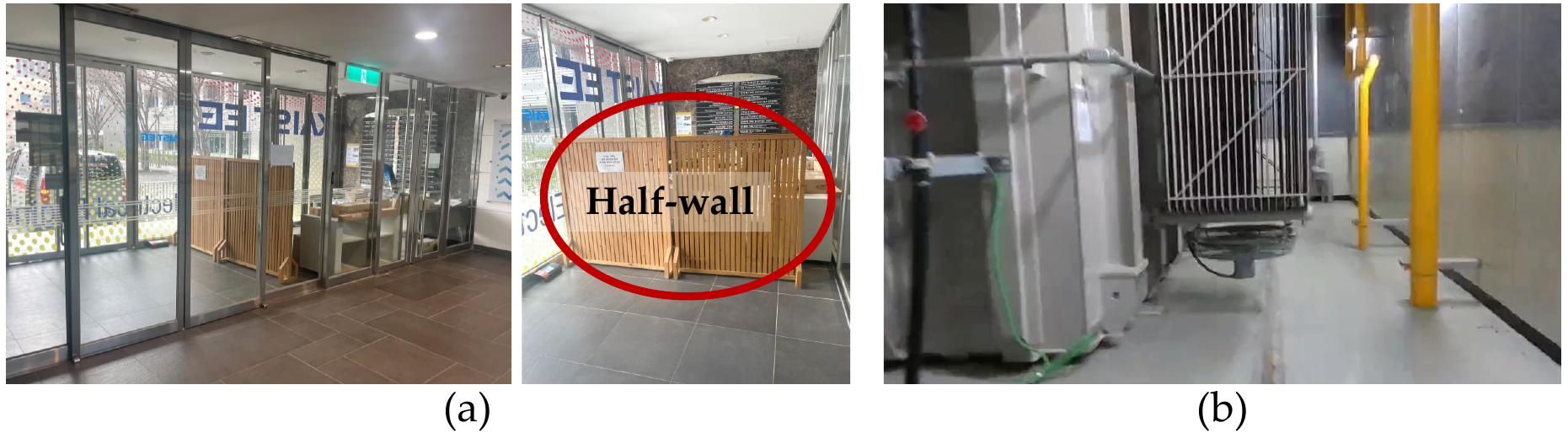}
    \caption{Real-world LiDAR dataset environments: (a)~{KAIST-E3} dataset with a half-wall separating the entrance and parcel storage areas. (b)~KEPCO dataset featuring large rooms with complex structures.}
    \label{fig:e3_kepco_env}
\end{figure}
We also collected handheld LiDAR datasets in real-world environments. The {KAIST-E3} dataset was recorded in the KAIST Electrical Engineering building, featuring a half-wall that restricts movement between the entrance and parcel storage areas~\mbox{(Fig.~\ref{fig:e3_kepco_env}(a))}. The KEPCO dataset was collected in a power station building with large rooms that contain complex wall-mounted structures and objects~\mbox{(Fig.~\ref{fig:e3_kepco_env}(b))}.

\definecolor{customRed}{rgb}{0.9,0.0,0.1}
\definecolor{customBlue}{rgb}{0.0,0.7,1.0}
\definecolor{customBrown}{rgb}{0.5,0.3,0.1}

\subsection{Consistency of Scene Graphs}
\label{sec:consistency_eval}
\subsubsection{Qualitative Evaluation}
\label{sec:consistency_qual}
\begin{figure}[t]
    \centering
    \includegraphics[width=0.9\linewidth]{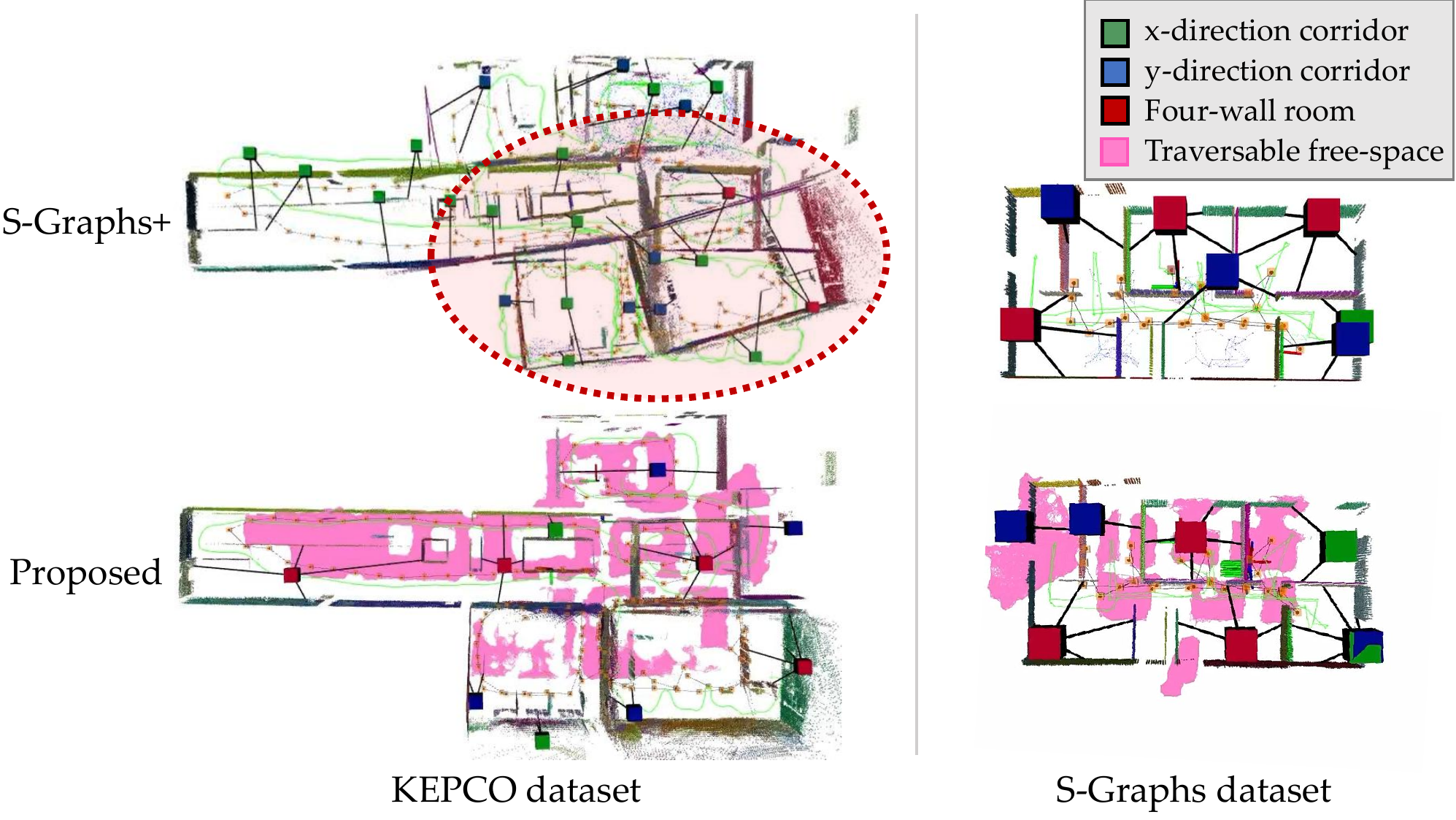}
    \caption{Comparison of room segmentation results between \mbox{S-Graphs+} and \mbox{\textit{TACS-Graphs}} across different environments. Both perform well on the simple S-Graphs dataset, but \mbox{S-Graphs+} exhibits over-segmentation and PGO distortion on the KEPCO dataset~(\textcolor{customRed}{red} dashed circle), while the proposed method mitigates both under- and over-segmentation.}
    \label{fig:sgraph_kepco}
\end{figure}

To demonstrate the effectiveness of the proposed method regardless of environmental complexity, experiments were conducted on both the simple S-Graphs dataset and the complex KEPCO dataset, comparing the performance of the \mbox{S-Graphs+} and \mbox{\textit{TACS-Graphs}}. The resulting scene graphs, which include both their nodes and the corresponding 3D points, are shown in \mbox{Fig.~\ref{fig:sgraph_kepco}}, where square markers represent rooms and lines indicate connections between walls and the corresponding rooms. Both methods perform well on the simple \mbox{S-Graphs} dataset. However, on the complex KEPCO dataset, \mbox{S-Graphs+} suffers from significant over-segmentation, generating an excessive number of room nodes. This disrupts PGO and distorts the 3D map geometry, as highlighted by the \textcolor{customRed}{red} dashed circle in the figure. The results indicate that excessive room segmentation negatively impacts PGO accuracy. In contrast, the proposed method effectively mitigates both under- and over-segmentation, achieving consistent room segmentation even in a complicated environment.

\begin{figure}[b]
    \centering
    \includegraphics[width=\linewidth]{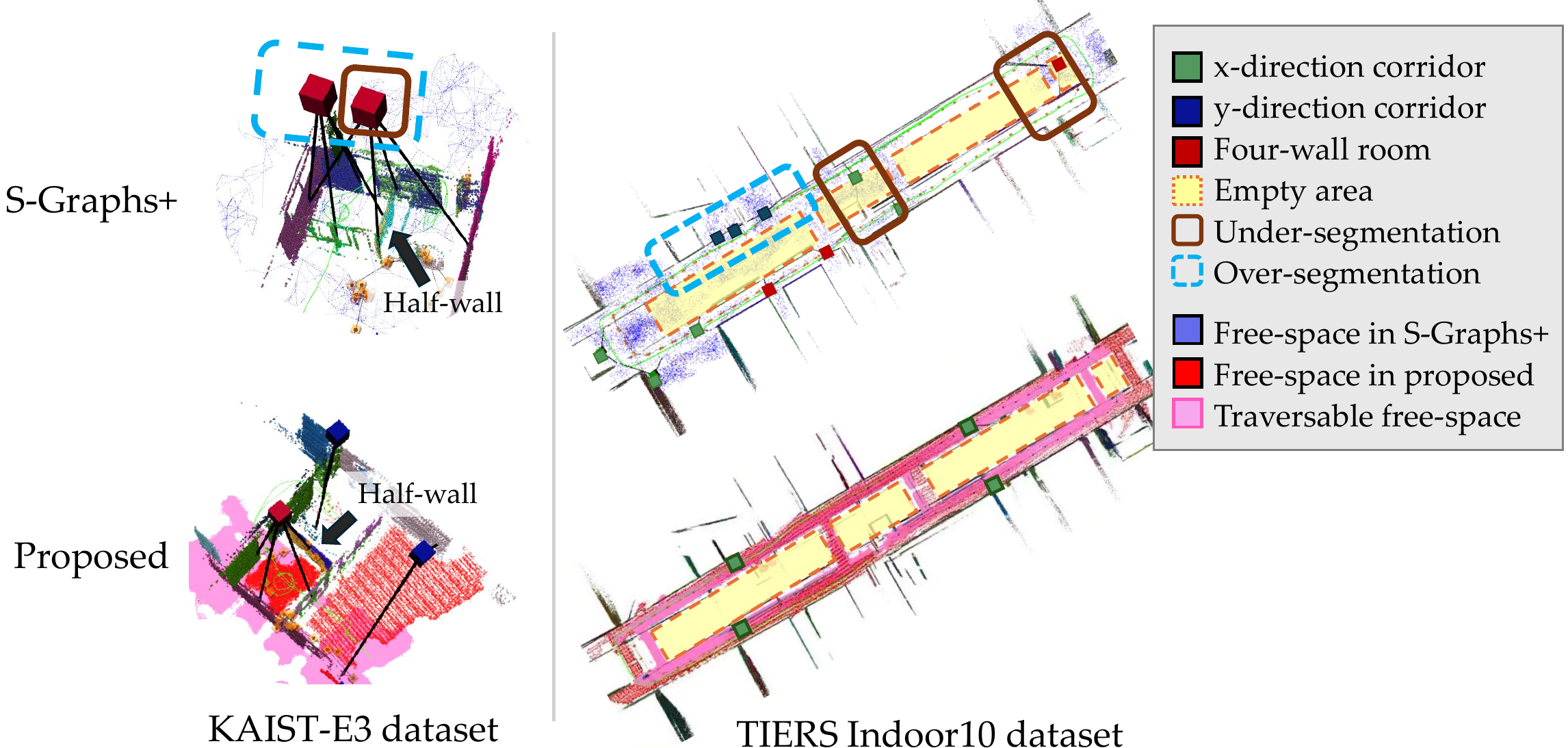}
    \caption{Room segmentation results using \mbox{S-Graphs+} and \mbox{\textit{TACS-Graphs}}. In \mbox{S-Graphs+}, the \textcolor{customBrown}{brown} boxes indicate under-segmentation of non-traversable room nodes, and \textcolor{customBlue}{blue} dashed boxes highlight over-segmentation. In contrast, \mbox{\textit{TACS-Graphs}} effectively segments semantically meaningful, traversable rooms.}
    \label{fig:e3_tiers10}
\end{figure}

Further experiments on the TIERS \texttt{Indoor10} and {KAIST-E3} datasets, which feature open spaces, confirm the effectiveness of the proposed method in addressing under-segmentation. As shown in \mbox{Fig.~\ref{fig:e3_tiers10}}, while \mbox{S-Graphs+} segments non-traversable areas as rooms, the proposed method successfully produces semantically meaningful, traversable segments. These results highlight its ability to prevent under-segmentation, especially in open environments.

\subsubsection{Quantitative Evaluation}
\label{sec:consistency_quan}
We quantitatively evaluated room segmentation consistency by comparing the proposed method with LiDAR-based scene graph approaches, including \mbox{S-Graphs+}~\cite{bavle2023s} and \mbox{GNN-based S-Graphs}~\cite{millan2023better}. Additionally, ablation studies were conducted with two \mbox{S-Graphs+} variants: one using a loop detection threshold of \mbox{$d^{\text{loop}}_{\text{th}}=5~\mathrm{m}$}, as in our method, to evaluate detection distance effects, and another incorporating the RM module to assess the impact of a traversable graph over voxel-based clustering. Scene graph consistency was measured using the following metrics:
\begin{itemize}
    \item Average number of rooms ($N_{\text{room}}$).
    \item Standard deviation of room counts ($\sigma_{\text{room}}$) and dice coefficient score~(DCS), quantifying the similarity of overlapping rooms in repeated trials.
    \item Total time spent on factor graph optimization ($t_{\text{FGO}}$).
\end{itemize}

DCS quantifies the similarity between two spatially close room regions \textit{A} and \textit{B} as follows:
\begin{equation}
\text{DCS} = 2 \cdot S(A \cap B) / (S(A) + S(B)),
\end{equation}
where $S(\cdot)$ denotes the area of the region. Higher DCS and lower $\sigma_{\text{room}}$ indicate better segmentation consistency, while lower $t_{\text{FGO}}$ reflects improved real-time performance through simplified and consistent graph representation.

\begin{table}[b!]
    \caption{Comparison of scene graph consistency on TIERS indoor datasets. Metrics include average room count~($N_{\mathrm{room}}$), standard deviation~($\sigma_{\mathrm{room}}$), dice coefficient score~(DCS), and factor graph optimization time~(\mbox{$t_{\mathrm{FGO}}$}).}
    \centering
    \scriptsize
    \setlength{\tabcolsep}{1pt}
    \begin{tabularx}{\columnwidth}{>{\centering\arraybackslash}m{0.8cm}|>{\centering\arraybackslash}m{2.8cm}||>{\centering\arraybackslash}X|>{\centering\arraybackslash}X|>{\centering\arraybackslash}X|>{\centering\arraybackslash}X }
    \hline
    \multirow{2}{*}{\textbf{Dataset}} 
    & \multirow{2}{*}{\textbf{Methods}}
    & \multicolumn{4}{c}{\textbf{Scene graph consistency}} \\ 
    \cline{3-6}
    & {} & \textbf{$N_{\mathrm{room}}$} & \textbf{$\sigma_{\mathrm{room}}$} $\downarrow$ & {DCS}(\%) $\uparrow$ & \textbf{$t_{\mathrm{FGO}}$}(s) $\downarrow$ \\ 
    \hline
    \multirow{5}{*}{\rotatebox[origin=c]{90}{Indoor07}} 
    & \raggedright \hspace{0.1em} {S-Graphs+~\cite{bavle2023s}} & 1.0 & 0.0 & 0.0 & \textbf{0.02} \\ 
    & \raggedright \hspace{0.1em} {GNN-based S-Graphs~\cite{millan2023better}} & 4.0 & 0.0 & 47.0 & 0.03 \\ 
    & \raggedright \hspace{0.1em} {S-Graphs+ with $d^{\text{loop}}_{\text{th}}$=$5\mathrm{m}$} & 1.0 & 0.0 & 0.0 & \textbf{0.02} \\ 
    & \raggedright \hspace{0.1em} {S-Graphs+ with RM} & 1.0 & 0.0 & 0.0 & 0.03 \\ 
    & \raggedright \hspace{0.1em} {Proposed} & 2.0 & 0.0 & \textbf{94.2} & \textbf{0.02} \\ 
    \hline
    \multirow{5}{*}{\rotatebox[origin=c]{90}{Indoor09}} 
    & \raggedright \hspace{0.1em} {S-Graphs+~\cite{bavle2023s}} & 5.7 & 1.1 & 10.7 & \textbf{0.09} \\ 
    & \raggedright \hspace{0.1em} {GNN-based S-Graphs~\cite{millan2023better}} & 11.7 & 8.01 & 22.9 & \textbf{0.09} \\ 
    & \raggedright \hspace{0.1em} {S-Graphs+ with $d^{\text{loop}}_{\text{th}}$=$5\mathrm{m}$} & 8.0 & 3.3 & 6.2 & 0.18 \\ 
    & \raggedright \hspace{0.1em} {S-Graphs+ with RM} & 17.5 & 1.5 & 18.1 & \textbf{0.09} \\ 
    & \raggedright \hspace{0.1em} {Proposed} & 2.67 & \textbf{0.94} & \textbf{46.5} & 0.82 \\ 
    \hline
    \multirow{5}{*}{\rotatebox[origin=c]{90}{Indoor10}} 
    & \raggedright \hspace{0.1em} {S-Graphs+~\cite{bavle2023s}} & 18.3 & 2.5 & 6.5 & 3.36 \\ 
    & \raggedright \hspace{0.1em} {GNN-based S-Graphs~\cite{millan2023better}} & 10.7 & 4.7 & 7.2 & 3.95 \\ 
    & \raggedright \hspace{0.1em} {S-Graphs+ with $d^{\text{loop}}_{\text{th}}$=$5\mathrm{m}$} & 10.0 & 2.1 & 3.2 & 0.93 \\ 
    & \raggedright \hspace{0.1em} {S-Graphs+ with RM} & 14.0 & 2.2 & 8.0 & 3.03 \\ 
    & \raggedright \hspace{0.1em} {Proposed} & 7.3 & \textbf{0.94} & \textbf{72} & \textbf{0.78} \\ 
    \hline
    \multirow{5}{*}{\rotatebox[origin=c]{90}{Indoor11}} 
    & \raggedright \hspace{0.1em} {S-Graphs+~\cite{bavle2023s}} & 16.7 & \textbf{0.47} & 17.3 & 0.46 \\ 
    & \raggedright \hspace{0.1em} {GNN-based S-Graphs~\cite{millan2023better}} & 12.2 & 8.9 & 14.5 & 0.41 \\ 
    & \raggedright \hspace{0.1em} {S-Graphs+ with $d^{\text{loop}}_{\text{th}}$=$5\mathrm{m}$} & 14.3 & 2.1 & 14.2 & 0.28 \\ 
    & \raggedright \hspace{0.1em} {S-Graphs+ with RM} & 19.5 & 2.5 & 32.6 & 0.55 \\ 
    & \raggedright \hspace{0.1em} {Proposed} & 14.0 & 1.4 & \textbf{35.3} & \textbf{0.26} \\ 
    \hline
    \end{tabularx}
    \label{tab:consistency_eval}
\end{table}

As shown in \mbox{Table~\ref{tab:consistency_eval}}, the proposed method outperforms all other approaches, achieving the highest DCS and lowest $\sigma_{\text{room}}$. However, in \texttt{Indoor11}, the performance of our method is slightly degraded due to the Manhattan world assumption, and the wide entrances in the environment cause the RM module to struggle. This reveals a limitation in handling unstructured environments, highlighting the need for further adaptation. Lower $N_{\text{room}}$ indicates reduced over-segmentation of \mbox{S-Graphs+}, and the corresponding reduction in $t_{\text{FGO}}$ suggests that the simplified graph representation enhances real-time performance. %The detailed analysis of these variants is described in \mbox{Section~\ref{sec:ablation}}. 
These results confirm the ability of \mbox{\textit{TACS-Graphs}} to effectively prevent under- and over-segmentation, thereby maintaining consistency in complex environments.

%%%%%%%%%%%%%%%%%%%%%%%%%%%%%%%%%%%%%%

\subsection{PGO Performance}
\label{sec:pgo_perf}
PGO performance through loop closure directly affects scene graph consistency, as inaccurate pose estimates lead to segmentation inconsistencies. Thus, efficient loop closure is crucial for maintaining graph consistency. PGO performance was evaluated using the same comparison methods and the TIERS dataset.

\subsubsection{Loop Closure Performance}
Loop closure performance was assessed using the following metrics: absolute trajectory error~(ATE), the number of loop detections~($N_{\text{loop}}$), the precision of detected loop pairs~($PR_{\text{loop}}$), which measures how accurately the actual positions of the keyframes align, and the total time spent on detecting loops during the entire process~(\mbox{$t_{\mathrm{LD}}$}). The results are summarized in \mbox{Table~\ref{tab:loop_eval}}.

\begin{table}[b]
    \caption{Comparison of loop closure performance on the TIERS indoor dataset. Metrics include ATE~(m), number of detected loops~($N_{\text{loop}}$), loop detection precision~($PR_{\text{loop}}$), and detection time~(\mbox{$t_{\mathrm{LD}}$}).}
    \centering
    \scriptsize
    \setlength{\tabcolsep}{1pt}
    \begin{tabularx}{\columnwidth}{>{\centering\arraybackslash}m{0.8cm}|>{\centering\arraybackslash}m{2.8cm}||>{\centering\arraybackslash}X|>{\centering\arraybackslash}X|>{\centering\arraybackslash}X|>{\centering\arraybackslash}X }
    \hline
    \multirow{2}{*}{\textbf{Dataset}} 
    & \multirow{2}{*}{{\textbf{Methods}}} 
    & \multicolumn{4}{c}{\textbf{Loop closure performance}} \\ 
    \cline{3-6}
    {} & {}
    & {ATE} (m) $\downarrow$
    & {$N_{\mathrm{loop}}$} 
    & {${PR}_{\mathrm{loop}}$} $\uparrow$
    & {$t_{\mathrm{LD}}$} (s) $\downarrow$ \\ 
    \hline
    \multirow{5}{*}{\rotatebox[origin=c]{90}{Indoor07}} 
    & \raggedright \hspace{0.1em} {S-Graphs+\cite{bavle2023s}} & \textbf{0.12} & 21 & 0.27 & 2.26 \\ 
    & \raggedright \hspace{0.1em} {S-Graphs+ with $d^{\text{loop}}_{\text{th}}$=$5\mathrm{m}$} & \textbf{0.12} & 21 & 0.28 & 2.26 \\ 
    & \raggedright \hspace{0.1em} {S-Graphs+ with RM} & 0.14 & 24 & 0.40 & 2.7 \\ 
    & \raggedright \hspace{0.1em} {GNN-based S-Graphs~\cite{millan2023better}} & 0.14 & 24 & 0.38 & 2.7 \\ 
    & \raggedright \hspace{0.1em} {Proposed}& 0.16 & 8.5 & \textbf{0.55} & \textbf{1.28} \\ 
    \hline
    \multirow{5}{*}{\rotatebox[origin=c]{90}{Indoor09}} 
    & \raggedright \hspace{0.1em} {S-Graphs+\cite{bavle2023s}} & 5.38 & 10.5 & 0.39 & 0.42 \\ 
    & \raggedright \hspace{0.1em} {S-Graphs+ with $d^{\text{loop}}_{\text{th}}$=$5\mathrm{m}$} & 23.06 & 56.6 & 0.29 & 14.45 \\ 
    & \raggedright \hspace{0.1em} {S-Graphs+ with RM} & 5.67 & 11.5 & 0.17 & 0.68 \\ 
    & \raggedright \hspace{0.1em} {GNN-based S-Graphs~\cite{millan2023better}} & 1.35 & 11.5 & 0.45 & 0.68 \\ 
    & \raggedright \hspace{0.1em} {Proposed} & \textbf{0.87} & 4.7 & \textbf{0.52} & \textbf{0.38} \\ 
    \hline
    \multirow{5}{*}{\rotatebox[origin=c]{90}{Indoor10}} 
    & \raggedright \hspace{0.1em} {S-Graphs+\cite{bavle2023s}} & {0.41} & 73.75 & 0.65 & 3.22 \\ 
    & \raggedright \hspace{0.1em} {S-Graphs+ with $d^{\text{loop}}_{\text{th}}$=$5\mathrm{m}$} & 0.42 & 137 & 0.35 & 23.71 \\ 
    & \raggedright \hspace{0.1em} {S-Graphs+ with RM} & 0.60 & 73.0 & 0.60 & 3.08 \\ 
    & \raggedright \hspace{0.1em} {GNN-based S-Graphs~\cite{millan2023better}} & 0.45 & 73.3 & \textbf{0.66} & 2.84 \\ 
    & \raggedright \hspace{0.1em} {Proposed}& \textbf{0.35} & 10.3 & \textbf{0.66} & \textbf{0.57} \\ 
    \hline
    \multirow{5}{*}{\rotatebox[origin=c]{90}{Indoor11}} 
    & \raggedright \hspace{0.1em} {S-Graphs+\cite{bavle2023s}} & \textbf{2.55} & 30.0 & 0.55 & 1.72 \\ 
    & \raggedright \hspace{0.1em} {S-Graphs+ with $d^{\text{loop}}_{\text{th}}$=$5\mathrm{m}$} & \textbf{2.55} & 108.0 & 0.08 & 21.15 \\ 
    & \raggedright \hspace{0.1em} {S-Graphs+ with RM} & \textbf{2.55} & 28.0 & \textbf{0.57} & {1.61} \\ 
    & \raggedright \hspace{0.1em} {GNN-based S-Graphs~\cite{millan2023better}} & \textbf{2.55} & 26.8 & 0.54 & \textbf{1.52} \\ 
    & \raggedright \hspace{0.1em} {Proposed} & 2.63 & 11.7 & 0.27 & 1.95 \\ 
     \hline
    \end{tabularx}
    \label{tab:loop_eval}
\end{table}

As existing scene graphs rely on repetitive loop closure detection over time, they inevitably exhibit higher $N_{\text{loop}}$ values. In contrast, \mbox{\textit{TACS-Graphs}}, with \mbox{\textit{CoSG-LCD}} module that selectively detects the most relevant loops when the same room is re-detected, achieves the highest $PR_{\text{loop}}$ despite detecting fewer loop pairs. This indicates that the loops detected are more likely to correspond to true matches. Notably, $t_{\text{LD}}$ is comparable to or shorter than other methods, demonstrating its efficiency.

\begin{figure}[b!]
    \centering 
    \includegraphics[width=0.8\linewidth]{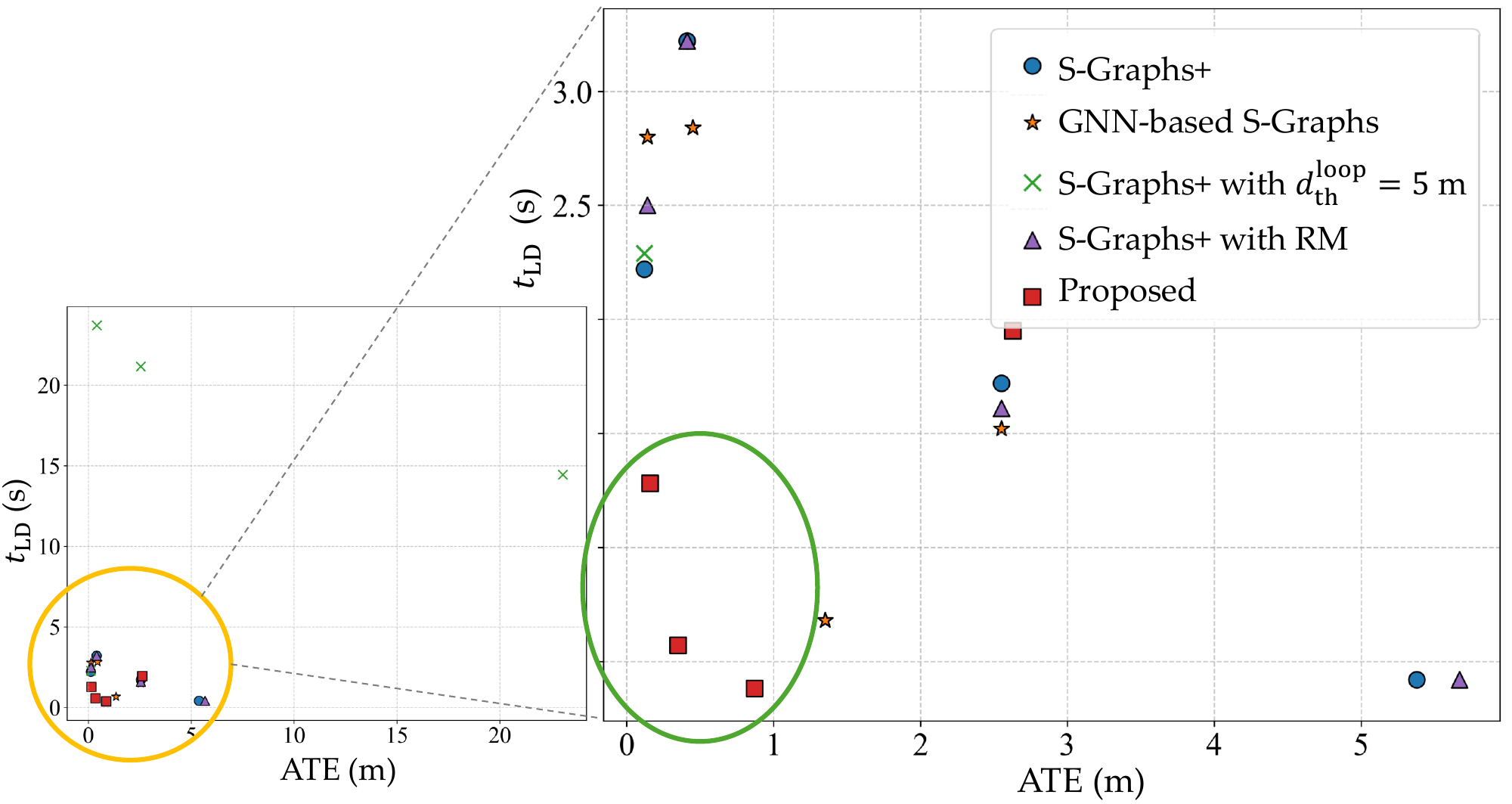}
    \caption{Comparison of loop detection efficiency. The \textit{x}-axis represents ATE (m), and the \textit{y}-axis indicates loop detection time ($t_{\text{LD}}$). The proposed method achieves lower ATE and faster detection than other approaches.} 
    \label{fig:scatter_whole} 
\end{figure}
\mbox{Fig.~\ref{fig:scatter_whole}} plots ATE versus loop detection time $t_{\text{LD}}$. As \mbox{S-Graphs+} with increased loop detection threshold~(\mbox{$d^{\text{loop}}_{\text{th}}=5~\mathrm{m}$}) resulted in significantly longer detection times and higher ATE, this variant was excluded, and the focus is placed on a closer examination near the origin. The graph shows that the proposed method achieves both lower ATE and faster loop detection than other approaches.

\subsubsection{Pose Estimation Accuracy}
Since loop closure aims to eliminate drift and refine the robot's pose, we evaluated pose estimation accuracy using ATE. However, to account for varying travel distances across datasets and to better assess performance on longer trajectories, we used the weighted absolute trajectory error (wATE) instead of standard ATE. The wATE is formulated as follows:
\begin{equation}
    \label{eq:wate}
    \text{wATE} = \frac{\sum_{i=1}^{N} D_i \cdot \text{ATE}_i}{\sum_{i=1}^{N} D_i},
\end{equation}
where $N$ is the total number of sequences, and $D_i$ represents the total trajectory length for each sequence.

We compared \mbox{\textit{TACS-Graphs}} with the followings: LiDAR-based SLAM methods incorporating scan context backends, \mbox{SC-ALOAM}~\cite{kim2022sclidarslam} and \mbox{ISC-LOAM}~\cite{wang2020iscloam}; Fast-GICP~\cite{koide2020fastgicp}, local odometry used for scene graph methods; and scene graph-based SLAM methods, \mbox{S-Graphs+} and \mbox{GNN-based S-Graphs}. 
As shown in \mbox{Table~\ref{tab:ate_comparison}}, \mbox{\textit{TACS-Graphs}} shows superior accuracy, achieving the lowest wATE. This improved localization accuracy reflects how enhanced scene graph consistency contributes to better loop closure and reduced drift. 

\begin{table}[t!]
    \caption{Comparison of pose estimation accuracy on the TIERS indoor dataset. Evaluated using ATE~(m) for each sequence and weighted ATE~(\mbox{wATE, m}) to account for varying travel distances across datasets. The travel distances for each sequence are $70.51~\mathrm{m}$, $152.25~\mathrm{m}$, $173.84~\mathrm{m}$, and $269.62~\mathrm{m}$, respectively.}
    \centering
    \scriptsize
    \setlength{\tabcolsep}{3.5pt} 
    \begin{tabularx}{\columnwidth}{>{\centering\arraybackslash}m{2.7cm}||>{\centering\arraybackslash}X|>{\centering\arraybackslash}X|>{\centering\arraybackslash}X|>{\centering\arraybackslash}X||>{\centering\arraybackslash}X}
    \hline
    \multirow{2}{*}{\diagbox{\textbf{Method}}{\textbf{Dataset}}} & \multicolumn{5}{c}{\textbf{ATE (m) $\downarrow$}} \\
    \cline{2-6}
    {} & {Indoor07} & {Indoor09} & {Indoor10} & {Indoor11} & {wATE}  \\
    \hline
    \raggedright SC-ALOAM~\cite{kim2022sclidarslam} & 0.36 & 5.27 & 1.81 & 9.92 & 5.73 \\ 
    \raggedright ISC-LOAM~\cite{wang2020iscloam} & 0.17 & 52.02 & 2.50 & 53.24 & 34.10 \\ 
    \hline
    \raggedright Fast-GICP~\cite{koide2020fastgicp} & 0.14 & 4.57 & 0.41 & 4.58 & 3.02 \\
    \raggedright S-Graphs+~\cite{bavle2023s} & \textbf{0.12} & 5.38 & 0.41 & \textbf{2.55} & 2.38 \\ 
    \raggedright GNN-based S-Graphs\cite{millan2023better} & 0.14 & 5.67 & 0.45 & \textbf{2.55} & 2.46 \\ 
    \raggedright Proposed & 0.16 & \textbf{0.87} & \textbf{0.35} & 2.63 & \textbf{1.37} \\ 
    \hline
    \end{tabularx}
\label{tab:ate_comparison}
\end{table}%These results validate the effectiveness of \textit{TACS-Graphs} in maintaining robust and efficient PGO performance in diverse environments.

\subsection{Ablation Studies}
\label{sec:ablation}
To validate the contribution of traversability to scene graph consistency, we analyzed two \mbox{S-Graphs+} variants while isolating the effects of loop detection distance threshold and the RM module: {S-Graphs+ with} \mbox{$d^{\text{loop}}_{\text{th}}=5~\textrm{m}$} and {S-Graphs+ with RM}. 
As shown in \mbox{Table~\ref{tab:loop_eval}}, the variant with \mbox{$d^{\text{loop}}_{\text{th}}=5~\mathrm{m}$} exhibited poor loop detection precision ($PR_{\text{loop}}$), leading to inaccurate loop closures and pose errors. These errors compromised the consistency of the scene graph, as reflected in \mbox{Table~\ref{tab:consistency_eval}}. 
Additionally, incorporating the RM module into \mbox{S-Graphs+}, which uses voxel-based free-space clustering, caused under-segmentation by merging rooms beyond their actual boundaries. This increased graph inconsistency, as evidenced by low DCS and high $\sigma_{\text{room}}$. 

These results underscore that inaccurate localization and inconsistent segmentation harm scene graph consistency and PGO performance. However, the proposed method's traversability-aware approach preserves graph structure and improves PGO, highlighting its importance in maintaining consistency regardless of loop closure threshold or RM module inclusion.
\section{Conclusion}
\label{sec:conclusion}

Existing 3D scene graph generation methods struggle with room segmentation inconsistencies, limiting their applicability in complex indoor environments. To address this, we proposed \mbox{\textit{TACS-Graphs}}, which incorporates a traversability map for ground robots to enhance scene graph consistency by mitigating under- and over-segmentation. Additionally, a novel loop closure detection module improves PGO efficiency and accuracy by leveraging scene graph consistency. Our key contribution is the identification of scene graph consistency as a critical challenge and its mitigation through traversability-based room segmentation. Extensive experiments in diverse environments validated our approach. However, the current approach is limited by the Manhattan world assumption, which restricts its applicability to rooms with rectilinear structures. As a result, it cannot be effectively applied to environments with irregular or non-orthogonal layouts. Future research will aim to develop more generalizable mapping methods that can handle a wide variety of indoor and outdoor environments, without relying on such structural assumptions.

\small
% \bibliographystyle{IEEEtran} %ieeenat_fullname}
% \bibliography{references}

\bibliographystyle{IEEEtran}
\bibliography{references}

% %%%%%%%%%%%%%%%%%%%%%%%%%%%%%%%%%%%%%%%%%%%%%%%%%%%%%%%%%%%%%%%%%%%%%%%%%%%%%%%%

% \section*{ACKNOWLEDGMENT}

\end{document}